\pdfoutput=1
\PassOptionsToPackage{dvipsnames}{xcolor}

\documentclass[11pt]{article}

\usepackage[]{EMNLP2022}

\usepackage{times}
\usepackage{latexsym}

\usepackage[export]{adjustbox}
\usepackage{verbatim}
\usepackage{graphicx}
\usepackage{multirow}
\usepackage{booktabs}
\usepackage[normalem]{ulem}
\useunder{\uline}{\ul}{}
\usepackage[all]{xy}
\usepackage{xcolor}
\usepackage{footmisc}
\usepackage{amsmath}
\usepackage{array}
\usepackage{lscape}
\usepackage{amssymb}
\usepackage{pifont}
\usepackage{tikz-cd}
\usepackage{epigraph}
\usepackage{enumitem} 

\usepackage[T1]{fontenc}

\usepackage[utf8]{inputenc}

\usepackage{microtype}

\usepackage{inconsolata}

\title{Learning Action-Effect Dynamics for \\Hypothetical Vision-Language Reasoning Task}

\author{Shailaja Keyur Sampat, Pratyay Banerjee, Yezhou Yang \and Chitta Baral \\
  Arizona State University, USA \\
  \texttt{\{ssampa17,pbanerj6,yz.yang,chitta\}@asu.edu} 
  }

\begin{document}
\maketitle
\begin{abstract}
`Actions' play a vital role in how humans interact with the world. Thus, autonomous agents that would assist us in everyday tasks also require the capability to perform `Reasoning about Actions \& Change' (RAC). This has been an important research direction in Artificial Intelligence (AI) in general, but the study of RAC with visual and linguistic inputs is relatively recent. The CLEVR\_HYP \cite{sampat2021clevr_hyp} is one such testbed for hypothetical vision-language reasoning with actions as the key focus. In this work, we propose a novel learning strategy that can improve reasoning about the effects of actions. We implement an encoder-decoder architecture to learn the representation of actions as vectors. We combine the aforementioned encoder-decoder architecture with existing modality parsers and a scene graph question answering model to evaluate our proposed system on the CLEVR\_HYP dataset. We conduct thorough experiments to demonstrate the effectiveness of our proposed approach and discuss its advantages over previous baselines in terms of performance, data efficiency, and generalization capability\footnote{Dataset setup scripts and code for baselines are 
available at \url{https://github.com/shailaja183/ARL}}.
\end{abstract}


\section{Introduction}

Humans interact with their environment to accomplish desired goals. Object manipulation (i.e., performing ``actions'' over the objects) is a fundamental concept that makes this interaction possible. In other words, actions in their simplest form have the power to change the state of a world and hence play a vital role in enabling humans to perform day-to-day tasks. As we are developing autonomous agents that can assist us in everyday tasks, they would also require to interact with complex environments. Hence, the development of autonomous agents that can perform actions to effectively manipulate objects and understand corresponding effects is of great importance. As a result, Reasoning about Action and Change (RAC) has been a long-standing research problem, since the rise of AI. 

  \begin{figure*}
\centering
\begin{minipage}{.48\linewidth}
\centering
  \includegraphics[width=\linewidth,left]{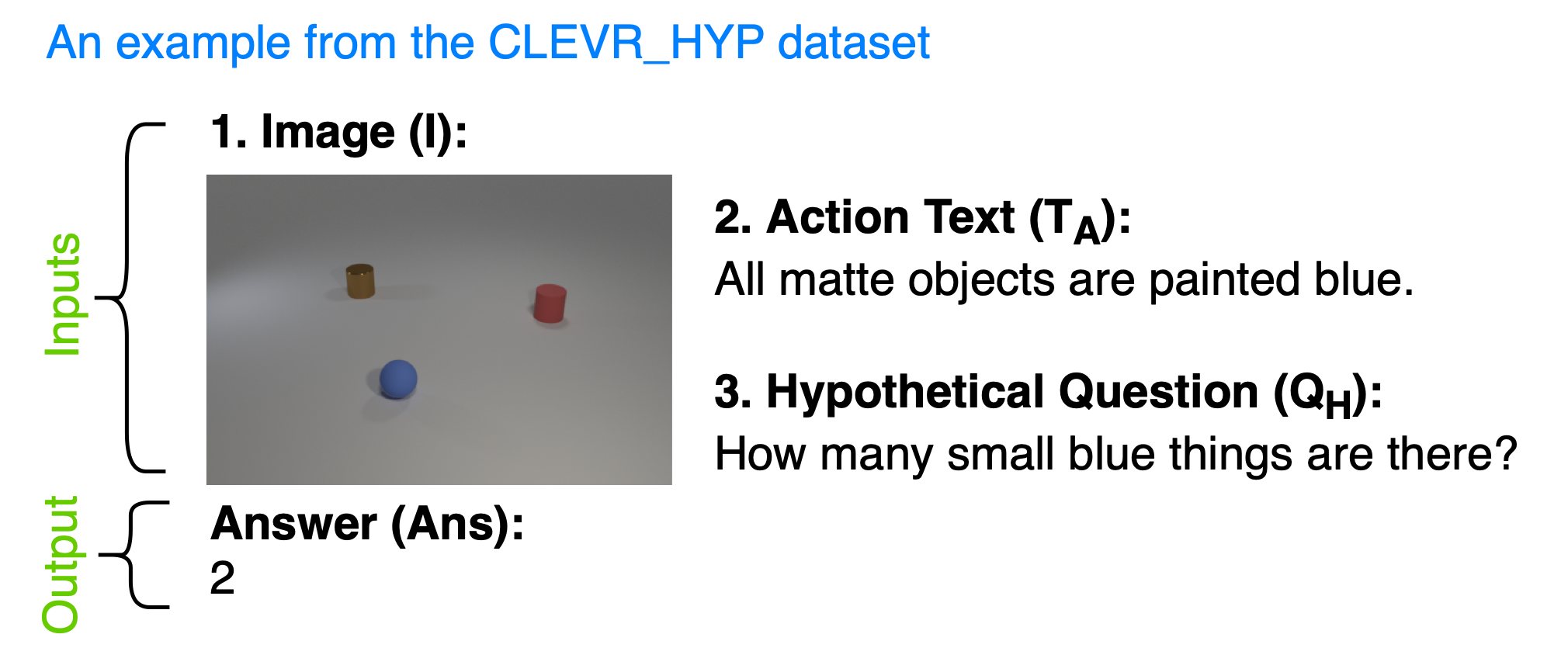}
  $ $
  \captionof{figure}{Revisiting CLEVR\_HYP task \cite{sampat2021clevr_hyp}: Answer a reasoning question (Q$_H$) about changes caused over the given image (I) by performing a hypothetical action (T$_A$).}
  \label{fig:open_example}
\end{minipage}
\hspace{0.3cm}
\begin{minipage}{.48\linewidth}
  \includegraphics[width=\linewidth]{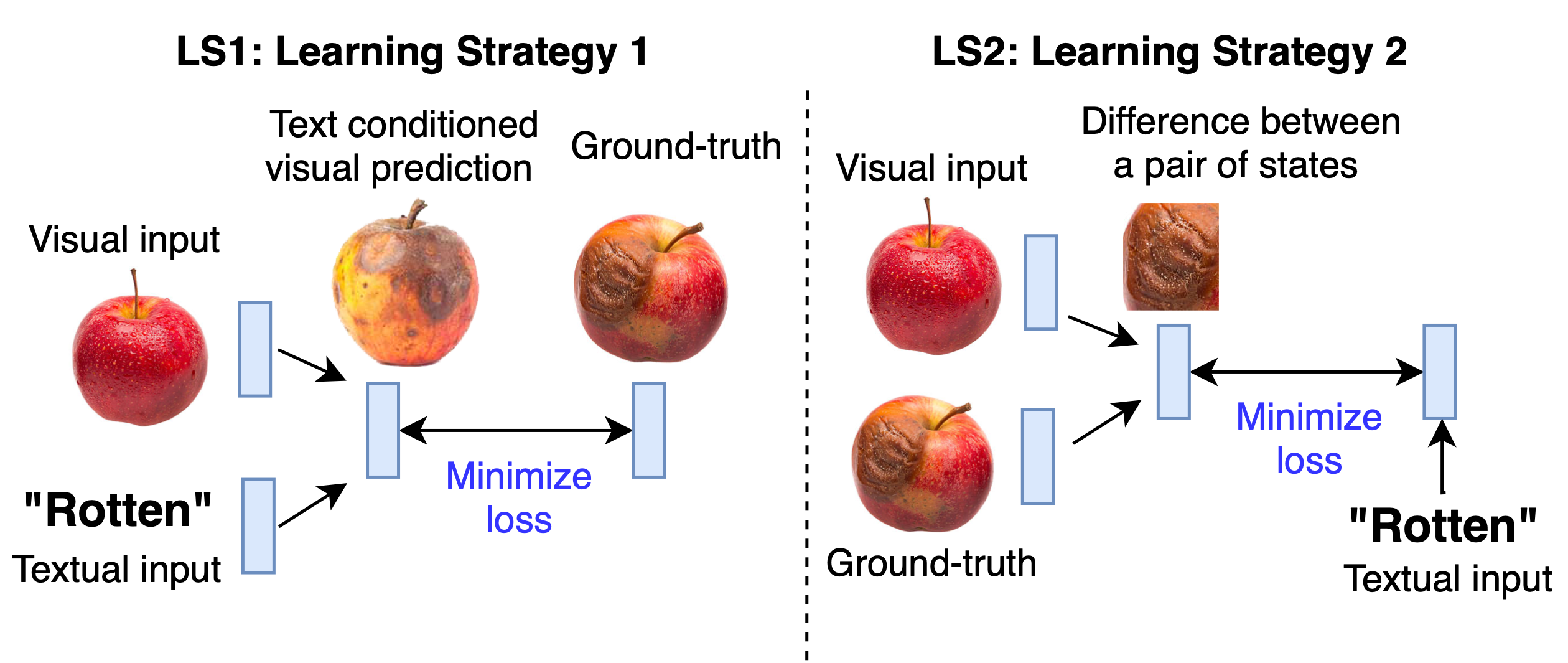}
  \captionof{figure}{{Two possible ways (LS1 and LS2) to learn action-effect dynamics in a supervised learning setting. In this paper, we implement LS2 which demonstrates improvements over LS1. Blue box denotes vector representation.}}
  \label{fig:intuition}
\end{minipage}%
\end{figure*}

The work of \citet{mccarthy1960programs} was the earliest to emphasize the importance of reasoning about actions. They developed an advice taker system that can do deductive reasoning about scenarios such as ``going to the airport from home'' requires ``walking to the car'' and ``driving the car to airport''. Since then, many real-life use cases have been identified which require AI models to understand interactions among the current states of the world, actions being performed over various objects, and most likely following states \cite{banerjee2020can}. While RAC has been more popular among knowledge representation and logic communities, it has recently piqued the interest of researchers in NLP and vision domains. A recent survey by  \citet{sampat2022reasoning} compiled a comprehensive list of  works that explore neural network's ability to reason about actions and changes, provided a dataset of linguistic and/or visual inputs.



In a recent tweet, Prof. Yann LeCunn also emphasized the importance of this research direction. He mentions that ``while we progress towards human-level AI, I believe we need to find new concepts that would (i) allow machines to learn to predict how one can influence the world through taking actions, (ii) learn hierarchical representations that allow long-term predictions in abstract spaces, (iii) enable agents to predict the effects of sequences of actions so as to be able to reason \& plan - \textit{all of this in ways that are compatible with gradient-based learning}'' \cite{ieee}. 

In this work, we aim to better tackle the hypothetical action reasoning task of CLEVR\_HYP dataset \cite{sampat2021clevr_hyp}. An example from this dataset is shown in Figure \ref{fig:open_example}). The key objective is to understand changes caused over a visual scene by an action described in the natural language and answer a reasoning question. In Figure \ref{fig:intuition}, we describe two possible action-effect learning strategies (LS1 and LS2) through a toy example to convey our intuition behind this work. LS1 uses visual features (i.e. features from the image of an apple) and a representation of actions (i.e. text ``rotten'') through sentence embeddings to imagine the effects (i.e. how a rotten apple would look like). This can be an intuitive choice to model the CLEVR\_HYP task using pre-trained vision-language models.

In our hypothesis, LS1 does not improve the model's understanding of what effects the actions will produce. Thus, we propose an alternative strategy, LS2\footnote{Figure \ref{fig:intuition} is meant to convey our intuition behind the proposed model in this work at a very high level. Figure \ref{fig:reasoner} is more complex in comparison and accurately describes the working of our model, considering the format of CLEVR\_HYP dataset, decomposing the task into various neural components, and measuring them using appropriate loss functions.}. Specifically, we let the model observe the difference between pairs of states before and after the action is performed (i.e. decayed portion of the apple that distinguishes a good apple from the rotten one), then associate those visual differences with the corresponding linguistic action descriptions (i.e. text ``rotten''). LS2 is likely to better capture action-effect dynamics, as action representations are learned explicitly.




To empirically test the above hypothesis, we develop a model (which is described in Section \ref{sec:arl}) and evaluate it on the CLEVR\_HYP dataset. We hope that our exciting results would enable the development of AI agents that can better collaborate with humans in the physical world and encourage further investigations in this research area. 
In summary, our key contributions are as follows;
\begin{itemize}[noitemsep,topsep=0pt,leftmargin=*]
    \item We propose a novel learning strategy for predicting the ``effects of actions'' in the vision-language domain (shown in Figure  \ref{fig:intuition}). 
    \item We develop a 3-stage model to implement the proposed learning strategy 
    and evaluate on the existing CLEVR\_HYP \cite{sampat2021clevr_hyp} dataset.
    \item Through ablations and analysis, we demonstrate the effectiveness of our model in terms of performance (5.9\% accuracy improvements), data efficiency (one-third of training data required), and better generalization capability in comparison with the best existing baselines. 
\end{itemize}

\begin{table*}
\centering
\small
\begin{tabular}{@{}lll@{}}
\toprule
\multicolumn{1}{c}{{ \textbf{Object Attributes in Visual Scenes}}}                                                         & \multicolumn{1}{c}{{ \textbf{Action Text Types}}} & \multicolumn{1}{c}{{ \textbf{Question Reasoning Types}}}                                                      \\ \midrule

1. \textbf{Shape}: cylinder, sphere or cube                                                                        & 1. \textbf{Add} new objects to the scene                       & 1.  \textbf{Counting} objects fulfilling the condition                                                                 \\
2. \textbf{Size}: small or big                                                                                     & 2. \textbf{Remove} objects from the scene                      & 2. \textbf{Verify existence} of certain objects                                                                        \\
3. \textbf{Material}: metal or rubber                                                                              & 3. \textbf{Change attribute} of the objects                   & 3. \textbf{Query attribute} of a particular object                                                                    \\
4. \textbf{Spatial}: left, right, front, behind or on 
 & 4. \textbf{Move} objects in or out of plane                         & 4. \textbf{Compare attributes} of two objects  \\
                                                                                                                              \begin{tabular}[c]{@{}l@{}} 5.  \textbf{Color}: red, green, gray, blue,\\ $\frac{}{} \frac{}{} \frac{}{}$ brown, yellow, purple or cyan\end{tabular}  &  
                                                                                                                              &     \begin{tabular}[c]{@{}l@{}} 5. \textbf{Integer comparison} of two object sets \\ $\frac{}{} \frac{}{} \frac{}{}$ (same, larger or smaller)\end{tabular}                                                                      \\ \bottomrule
\end{tabular}
\caption{Summary of object attributes, actions, and reasoning types in CLEVR\_HYP dataset \cite{sampat2021clevr_hyp}}
\label{tab:taxonomy}
\end{table*} 

\section{CLEVR\_HYP}
\label{sec:clevr_hyp}  
In this section, we briefly summarize important aspects of the CLEVR\_HYP dataset \cite{sampat2021clevr_hyp} and related terminologies used in the subsequent sections. 

\subsection{Problem Formulation} 
The task aims at understanding changes caused over an image by performing an action described in natural language and then answering a reasoning question over the resulting scene. Consult Figure \ref{fig:open_example} for better understanding of the following;

\begin{itemize}[noitemsep,topsep=1pt,leftmargin=*]
\item \textit{Inputs:} 
\begin{enumerate}[noitemsep,topsep=0pt,leftmargin=*]
    \item Image (I)- Visual scene with rendered objects
    \item Action Text (T$_A$)- Textual modality describing action(s) to be performed over I
    \item Hypothetical Question (Q$_H$)- Textual question that will assess the system's capability to understand changes caused by T$_A$ on I
\end{enumerate}
\item \textit{Output:} Answer (A) for the given Q$_H$
\item \textit{Answer Vocabulary:} [0-9, yes, no, cylinder, sphere, cube, small, big, metal, rubber, red, green, gray, blue, brown, yellow, purple, cyan]
\item \textit{Evaluation:} 27-class Answer Classification / Accuracy (\%)
\end{itemize}

\subsection{Dataset Details and Partitions}
\label{sec:datapart}
The CLEVR\_HYP dataset assumes to have a closed set of object attributes, action types, and question reasoning types which are summarized in Table \ref{tab:taxonomy}. The dataset is divided into the following partitions;

\begin{itemize}[noitemsep,topsep=1pt,leftmargin=*]
    \item \textit{Train} (67.5k) /  \textit{Val} (13.5k) sets have <I, T$_A$, Q$_H$, A> tuples along with the scene graphs as a visual oracle and functional programs\footnote{Originally introduced in CLEVR \cite{johnson2017clevr}. For example, a question `How many red metal things are there?' can be represented as a functional program `count(filter\_color(filter\_material(scene(),metal),red))'} as a textual oracle.  
    \item \textit{Test} sets consist of only <I, T$_A$, Q$_H$, A> tuples, and \textit{no oracle annotations are available}. There are three different test sets,
    \begin{enumerate}[noitemsep,topsep=0pt,leftmargin=*]
    \item Ordinary test (13.5k) consists of examples with the same difficulty as train/val
    \item 2HopT$_A$ test (1.5k) consists of examples where two actions are performed ex. `Move a purple object on a red cube \textit{then} paint it cyan.'
    \item 2HopQ$_H$ test (1.5k) consists of examples where two reasoning types are combined ex. `How many objects are \textit{either} red \textit{or} cylinder?'
    \end{enumerate}
\end{itemize}

\subsection{\textbf{Baseline Models}}
\label{sec:bl}
Following is a brief description of two top-performing baselines reported in \citet{sampat2021clevr_hyp}, to which we will compare the results of our proposed approach in this paper. 

\begin{itemize}[noitemsep,topsep=3pt,leftmargin=*]
    \item \textbf{(TIE) Text-conditioned Image Editing:} Text-adaptive encoder-decoder with residual gating \cite{vo2019composing} is used to generate new image conditioned on the action. Then, new image along with the question is fed into LXMERT \cite{tan2019lxmert} (which is a pre-trained vision-language transformer), to generate an answer. The model can be visualized in Figure \ref{fig:bl2}. 
    
    \begin{figure}[ht!]
\centering
  \includegraphics[width=0.9\linewidth]{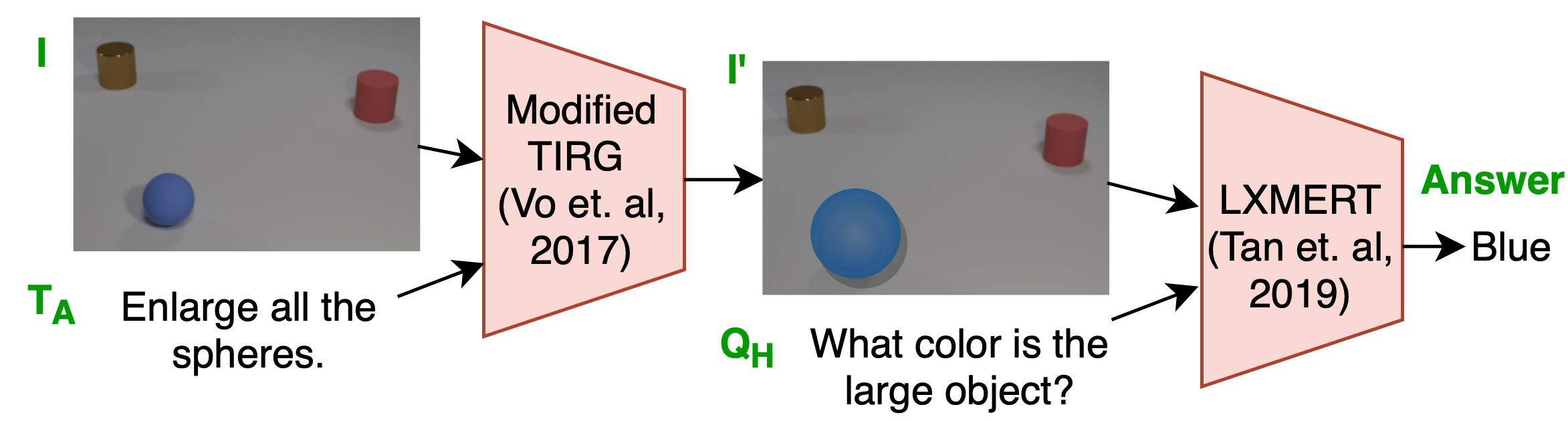} 
  \caption{Architecture of TIE baseline}
  \label{fig:bl2}
\end{figure}
    
    \item \textbf{(SGU) Scene Graph Update:} In this model, understanding changes caused by an action text is considered as a graph-editing problem. First, an image is converted into a scene graph and action text is converted into a functional program (FP). \citet{sampat2021clevr_hyp} developed a module inspired by \citet{chen2020graph} that can generate an updated scene graph based on the original scene graph and a functional program of an action text. It is followed by a neural-symbolic VQA model \cite{yi2018neural} that can generate an answer to the question provided the updated scene graph. The model can be visualized in Figure \ref{fig:bl4}.
    \begin{figure}[ht!]
\centering
  \includegraphics[width=0.9\linewidth]{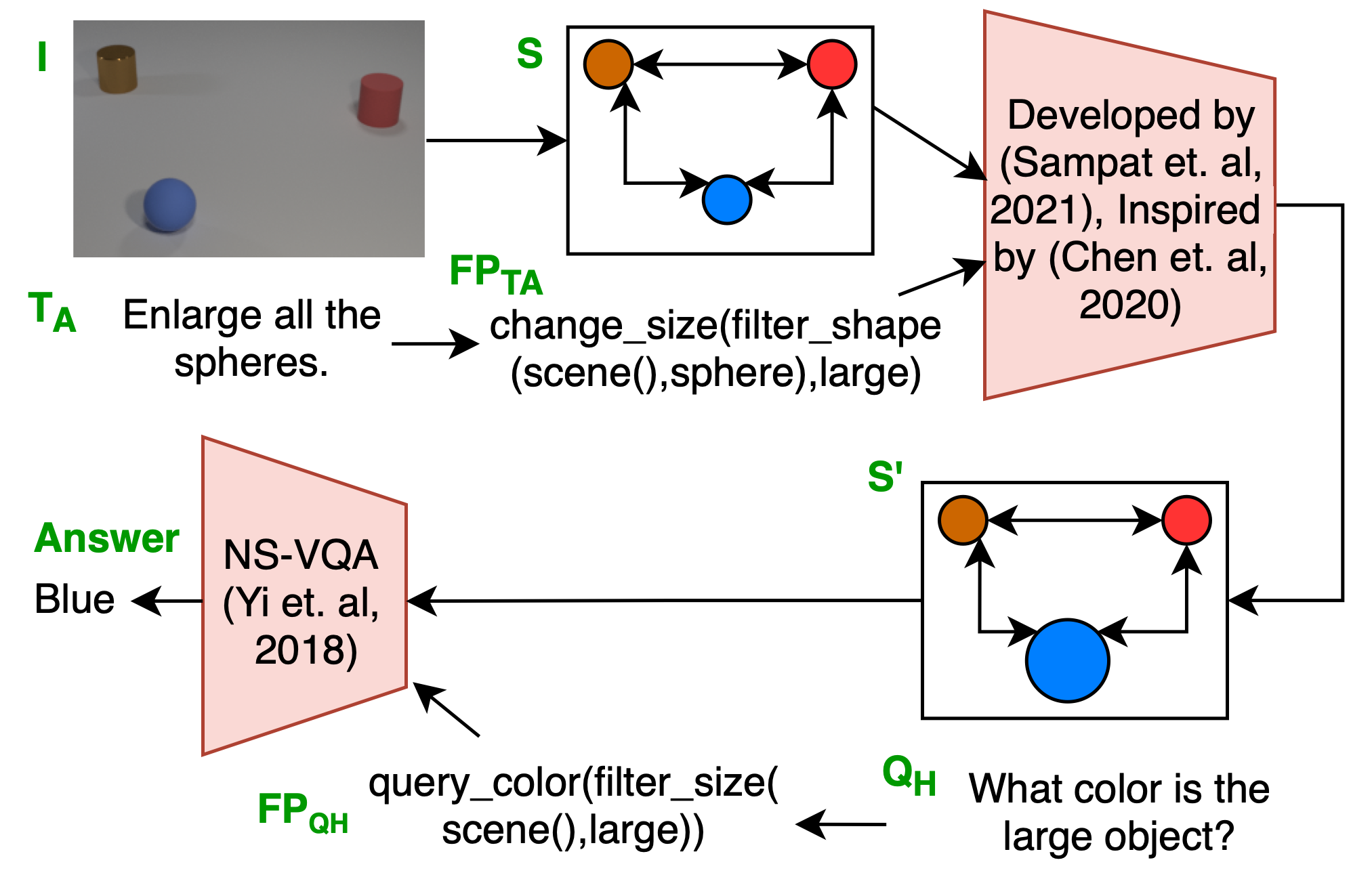} 
  \caption{Architecture of SGU baseline}
  \label{fig:bl4}
\end{figure}
\end{itemize}

\begin{figure*}
    \centering
  \includegraphics[width=\linewidth]{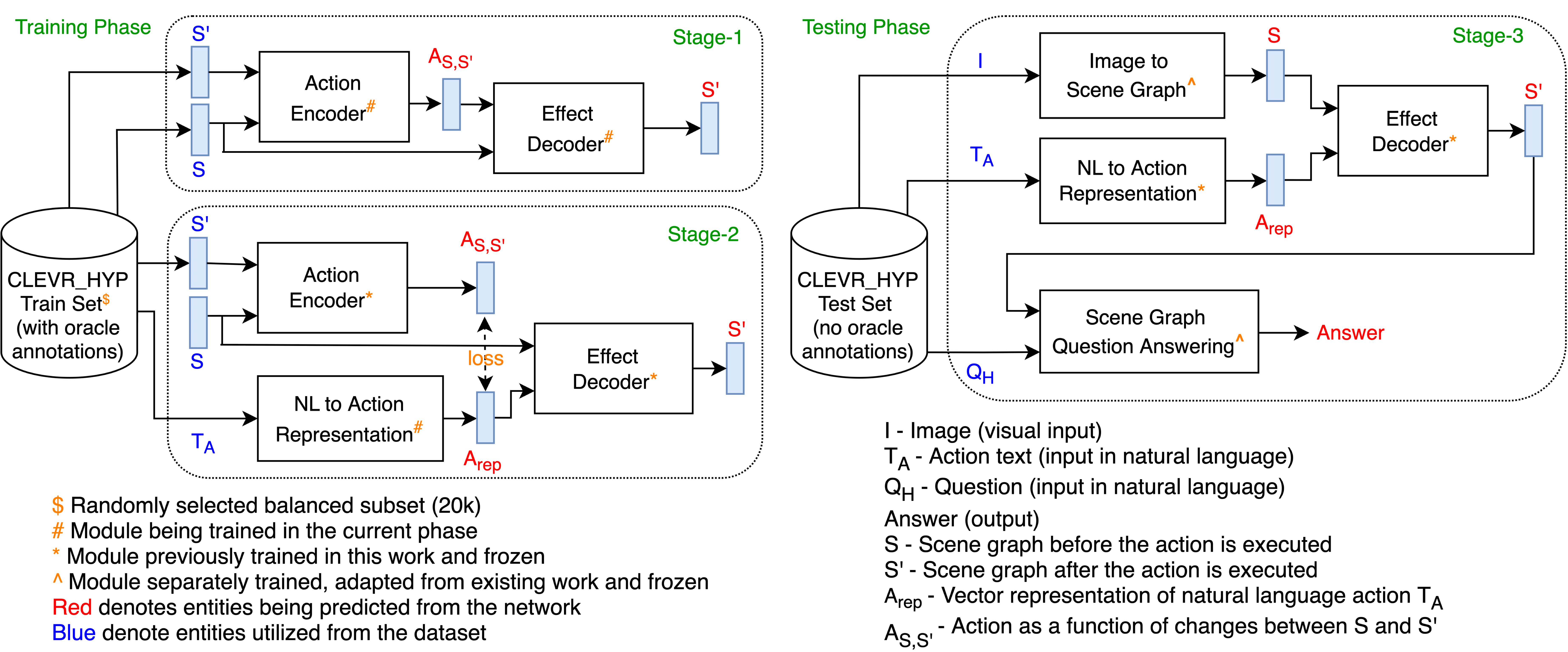} 
  \caption{Detailed visualization of our proposed 3-stage Action Representation Learner (ARL) model: (left) training phase (right) testing phase (bottom) terms and notations used. Best viewed in color. }
  \label{fig:reasoner}
\end{figure*}

There are important distinctions between baselines developed by \citet{sampat2021clevr_hyp} and our proposed method. Both the above baselines rely on pre-learned word representation of actions- either by a word-vector algorithm or a learned functional program and use that to conditionally update the visual scene (at pixel level or through graph operations). Thus, TIE and SGU resembles more to LS1 in Figure \ref{fig:intuition}. Note that in SGU baseline, individual functions (in their functional program representations) are human authored i.e. what kind of inputs it accepts and what it will return when executed. For example, ‘remove $<$attribute$>$’ function will take a set of objects as input and return a subset of objects which do not have $<$attribute$>$. 

In contrast, we learn action representations through two-step process. First, we learn to predict changes in a pair of scene-graphs (before and after the action is performed). And second, we minimize the loss between changes in the scene with the representation of linguistic action descriptions. Thus, our proposed model resembles more to LS2 in Figure \ref{fig:intuition}. Our method is purely based on data and does not require any human intervention. Also, note that oracle scene graphs post-actions are not available at the test-time. By enforcing two-way representation learning, we are able to predict changes in the scene graph using and action vector from linguistic action description for a given test instance.

\section{Proposed Model: Action Representation Learner (ARL)}
\label{sec:arl}

In this section, we describe the architecture of our proposed model Action Representation Learner (ARL). 

\begin{figure}
    \centering
  \includegraphics[width=\linewidth]{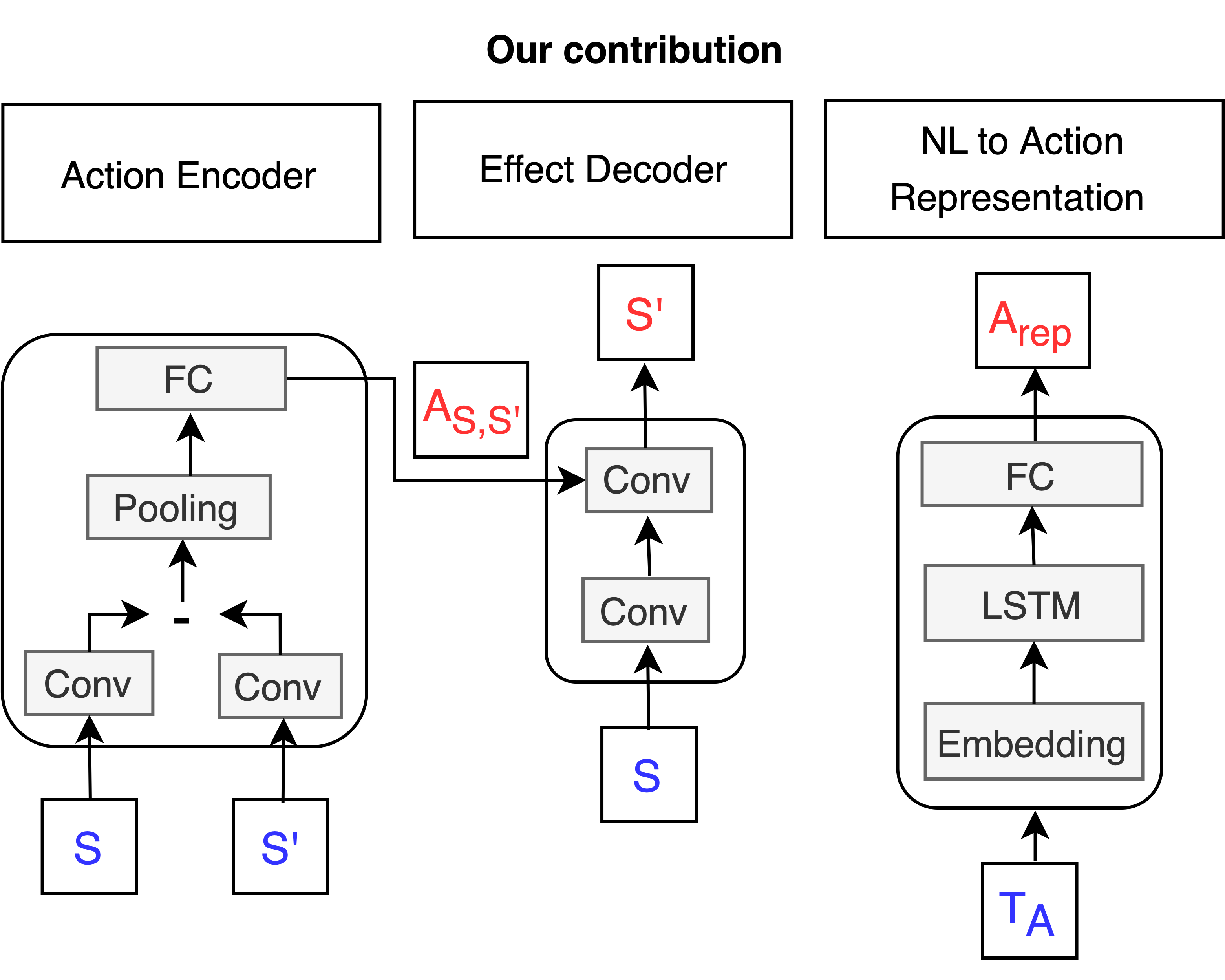} 
  \caption{Internal details of the components used in ARL model. \textcolor{red}{Red} denotes entities predicted, \textcolor{blue}{Blue} denotes entities utilized from the dataset.}
  \label{fig:internal}
\end{figure}

In our point of view, the most critical component of a model that attempts to solve CLEVR\_HYP is the one where mapping between visual changes and actions are learned. In Figure \ref{fig:intuition}, we graphically demonstrated our intuition behind how we can do so. Our hypothesis is that a model can learn better action representations by observing difference between a pair of states (before and after the action is performed) and then associate those visual differences with given linguistic description of actions. In this regard, we attempt to create a 3-stage model shown in Figure \ref{fig:reasoner}, which we believe would better capture the causal structure of this task. Detailed description of each individual component is provided below and can be visualized in Figure \ref{fig:internal}.

\subsection{Stage-1}
Actions have the power to change the state of the world. In other words, difference between a pair of states can be considered as a function of performing actions. For example, consider two states `green cube' and `red cube'. The change between above states is `green$\rightarrow$red', which is a function of action `paint' or `change color'. Conversely, if one is provided with a state `green cube' and knowing that the `paint red' action is performed, then one can visualize that the next state would be `red cube'. We aim to capture such relationship between state changes and actions in this stage. 

Specifically, we setup an encoder-decoder model to achieve this objective. Since our objective is to learn action and effects, we refer to them as `Action Encoder' and `Effect Decoder'. As described in Section \ref{sec:datapart}, the training set of CLEVR\_HYP provides oracle annotations for an initial scene graph S (using which the image is rendered) and a scene graph after executing the action text S$^{\prime}$. We take a random subset of 20k\footnote{We experiment with different data sizes and discuss results in Section \ref{sec:abl}, but obtain the optimal results for 20k samples when action vector length is 125} scene graph pairs from CLEVR\_HYP training set that are balanced by action types (add, remove, change, move) to train this encoder-decoder. 

We capture the difference between states S and S$^{\prime}$ i.e. A$_{S,S'}$ using the encoder. At the test time, we do not have the updated scene graph S$^{\prime}$ available. To address this issue, the encoder is followed by a decoder, which can reconstruct S$^{\prime}$ provided S and the learned scene difference A$_{S,S'}$ in the encoder network. Formally, 

\begin{equation}
\textcolor{red}{A_{S,S^{\prime}}} = Action Encoder(\textcolor{blue}{S}, \textcolor{blue}{S^{\prime}})  
\end{equation}

\begin{equation}
\textcolor{red}{S^{\prime}} = Effect Decoder(\textcolor{blue}{S}, \textcolor{red}{A_{S,S^{\prime}}})
\end{equation}

Where $(\textcolor{blue}{S}, \textcolor{blue}{S'}) \in$ CLEVR\_HYP training set,
\textcolor{red}{red} denotes entities predicted, \textcolor{blue}{blue} denotes entities utilized from the dataset. We jointly train action encoder-effect decoder networks with the following objective;
\begin{multline}
argmax_{\Theta _{ActionEncoder}\Theta _{EffectDecoder}}^{} \\  [ log P(\textcolor{red}{S^{\prime}}|\frac{}{}\textcolor{blue}{S},\frac{}{}ActionEncoder(\textcolor{blue}{S}, \textcolor{blue}{S^{\prime}})) ]
\end{multline}

Note that, in common applications involving encoder-decoder architecture, the decoder part of the model is removed once the desired performance is achieved and the encoder is used to encode input sequences to a fixed-length vector at test-time. Contrary, here we discard the encoder part and keep the decoder part to obtain updated scene representation from the initial scene and learned action vector. 

\subsection{Stage-2}
\label{sec:parser}

As explained in stage-1, we do not have the S$^{\prime}$ at the test time and we cannot compute A$_{S,S'}$. However, provided that changes in the scene are a function of the action, we can approximate A$_{rep}$ A$_{rep}$ is a vector representation corresponding to natural language action. A network is trained which can convert `Natural language to Action Representation' with the help of encoder-decoder network trained in Stage-1 that maximizes the log probability that outputs the correct state S$^{\prime}$ as below. 

 \begin{multline}
argmax_{\Theta _{NL2ActionRep}}^{} \\  [ log P(\textcolor{red}{S^{\prime}}|\frac{}{}\textcolor{blue}{S},\frac{}{}NL2ActionRep(\textcolor{blue}{T_A})) ]
\end{multline}

At the core, lies LSTM encoder, which precedes by an embedding layer and followed by dense layers. During model training, in addition to finding the values for the weights of the LSTM and dense layers, the word embeddings for each word in the training set are computed. This is achieved using nn.Embedding(vocabulary\_size, embedding\_size) layer defined in pytorch. 
 This way, a fixed length one-hot vector of given length is generated for each word in the vocabulary depending on the position of the word in context and  updated using back-propagation. Embedding layer is similar to a linear layer, which returns the index where one is located instead of returning the whole one-hot vector. It takes an action text T$_A$ as a sequence of learned word embeddings, runs an LSTM over them, then projects from the final cell state to get the output A$_{rep}$. The LSTM has a hidden layer of size 200.

\subsection{Stage-3}
\label{sec:stage3}

For each image, we use Mask R-CNN \cite{he2017mask} to generate segment proposals of all objects. Along with the segmentation mask, the network also classifies the objects based on their visual attributes- color, material, size, and shape. The threshold for segment proposals is set to 0.9 i.e. segments with bounding-box score less than 0.9 are dropped. The segment for each single object, paired with the original image (resized to 224x224) is sent to ResNet-34 \cite{he2016deep} to extract 3D coordinates of objects in the scene. Inclusion of original full image is observed to enhance the performance by incorporating contextual information. Note that scene-parsing is pre-trained and not fine-tuned with rest of the network. 


\section{Results and Analysis}

In this section, we discuss the performance of our model quantitatively and qualitatively. Additionally, we discuss our findings from three ablations for our model.

\subsection{Quantitative Results}

\noindent Once we complete the aforementioned 3-stage training process, we leverage a couple of existing models along with the trained components to make predictions on CLEVR\_HYP \cite{sampat2021clevr_hyp} test data (as shown in the right part of the Figure \ref{fig:reasoner}). The CLEVR\_HYP has three test sets- Ordinary, 2HopT$_A$ and 2HopQ$_H$. Refer to Section \ref{sec:datapart} for the description of each test setting with examples.    

\begin{table}[ht!]
\centering
\begin{tabular}{llccc}
\hline
\multicolumn{5}{c}{\textbf{Test performance on CLEVR\_HYP}} \\ \hline
  &  & \textit{TIE} & \textit{SGU} & \textit{ARL} \\
\textit{Ordinary} &  & 64.7 & 70.5 & 76.4 \\
\textit{2HopA$_T$} &  & 55.6 & 64.4 & 69.2 \\
\textit{2HopQ$_H$} &  & 58.7 & 66.5 & 70.7 \\ \hline
\end{tabular}
\caption{Performance of two baselines (TIE, SGU) reported in \cite{sampat2021clevr_hyp} and our proposed model (ARL) on three test sets of CLEVR\_HYP}
\label{tab:quanr}
\end{table}

\begin{table}[h]
\centering
\begin{tabular}{@{}llllc@{}}
\toprule
\multicolumn{5}{c}{\textbf{Accuracy(\%) by Action Types}} \\ \midrule
\multicolumn{1}{c}{{\ul \textit{Validation}}} &  & TIE & SGU & \multicolumn{1}{l}{$\frac{}{}$ ARL} \\
Add &  & \multicolumn{1}{c}{58.2} & \multicolumn{1}{c}{65.9} & 70.3 \\
Remove &  & 89.4 & 88.6 & 94.1 \\
Change &  & 88.7 & 91.2 & 95.8 \\
Move &  & 61.5 & 69.4 & 72.6 \\ \midrule
\multicolumn{1}{c}{{\ul \textit{2HopT$_A$}}} &  & \textit{TIE} & \textit{SGU} & \multicolumn{1}{l}{$\frac{}{}$ ARL} \\
Add + Remove &  & \multicolumn{1}{c}{53.6} & \multicolumn{1}{c}{63.2} & 66.7 \\
Add + Change &  & 55.4 & 64.7 & 70.6 \\
Add + Move &  & 49.7 & 57.5 & 63.2 \\
Remove + Change &  & 82.1 & 85.5 & 91.6\\
Remove + Move &  & 52.6 & 66.4 & 68.3 \\
Change + Move &  & 53.8 & 63.3 & 67.1 \\ \bottomrule
\end{tabular}
\caption{Performance breakdown of models by different action types in validation and 2HopT$_A$ test set}
\label{tab:actype}

\end{table}

\begin{table}[ht!]
\centering
\begin{tabular}{llccc}
\hline
\multicolumn{5}{c}{\textbf{Accuracy(\%) by Reasoning Types}} \\ \hline
\multicolumn{1}{c}{{\ul \textit{Validation}}} &  & TIE & SGU & ARL \\
Count &  & 60.2 & 74.3 & 78.6 \\
Exist &  & 69.6 & 72.6 & 77.3 \\
CompareInteger &  & 56.7 & 67.3 & 70.7 \\
CompareAttribute &  & 68.7 & 70.5 & 73.4 \\
QueryAttribute &  & 65.4 & 68.1 & 74.9 \\ \hline
\multicolumn{1}{c}{{\ul \textit{2HopQ$_H$}}} &  & \textit{TIE} & \textit{SGU} & \textit{ARL} \\
And &  & 59.2 & 67.1 & 70.3 \\
Or &  & 58.8 & 67.4 & 71.5 \\
Not &  & 58.1 & 65.0 & 68.4 \\ \hline
\end{tabular}
\caption{Performance breakdown of models by different question types in validation and 2HopQ$_H$ test set}
\label{tab:qhtype}
\end{table}

The CLEVR\_HYP dataset is formulated as a classification task with exactly one correct answer. Therefore, the exact match accuracy (\%) metric is used for evaluation. Table \ref{tab:quanr} demonstrates the performance of our proposed model in comparison with the two best performing existing models TIE and SGU, described in Section \ref{sec:bl}. Our proposed approach outperforms those baselines by 5.9\%, 4.8\% and 4.2\% on \textit{Ordinary}, \textit{2HopT$_A$ Test} and \textit{2HopQ$_H$ Test} respectively. This demonstrates that our model not only achieves better overall accuracy but also has improved generalization capability when multiple actions have to be performed on the image or understand logical combinations of attributes while performing reasoning. 

In Table \ref{tab:actype}, we analyze the ability of models to perform a particular action. For validation set, the model is expected to perform one of the four actions- add objects, remove objects, change in attributes or move objects. Overall, we can observe that our proposed method ARL achieves better fine-grained accuracy for all action types compared to existing models. All three models do quite well on `remove' and `change' actions whereas struggle when new objects are added or existing objects are moved around. Yet, our model shows 4.4\% and 3.2\% improvements on `add' and `move' respectively compared to the best previous baseline. 

For 2HopT$_A$ test set, the model is expected to perform two different actions (among add, remove, change and move) one after the other. Our observation from the Validation results remains consistent when multiple actions are combined. In other words, models were able to achieve relatively high accuracy for actions `remove' and `change', hence their combination `remove+change' also has high model performance. Whereas other combinations of actions accomplish relatively lower performance. It leads to the conclusion that understanding the effect of different actions is of varying complexity. It also indicates that the learned action representations in our proposed model are helpful as it shows better generalization by action types. 

\begin{figure*}
\centering
  \includegraphics[width=0.85\linewidth]{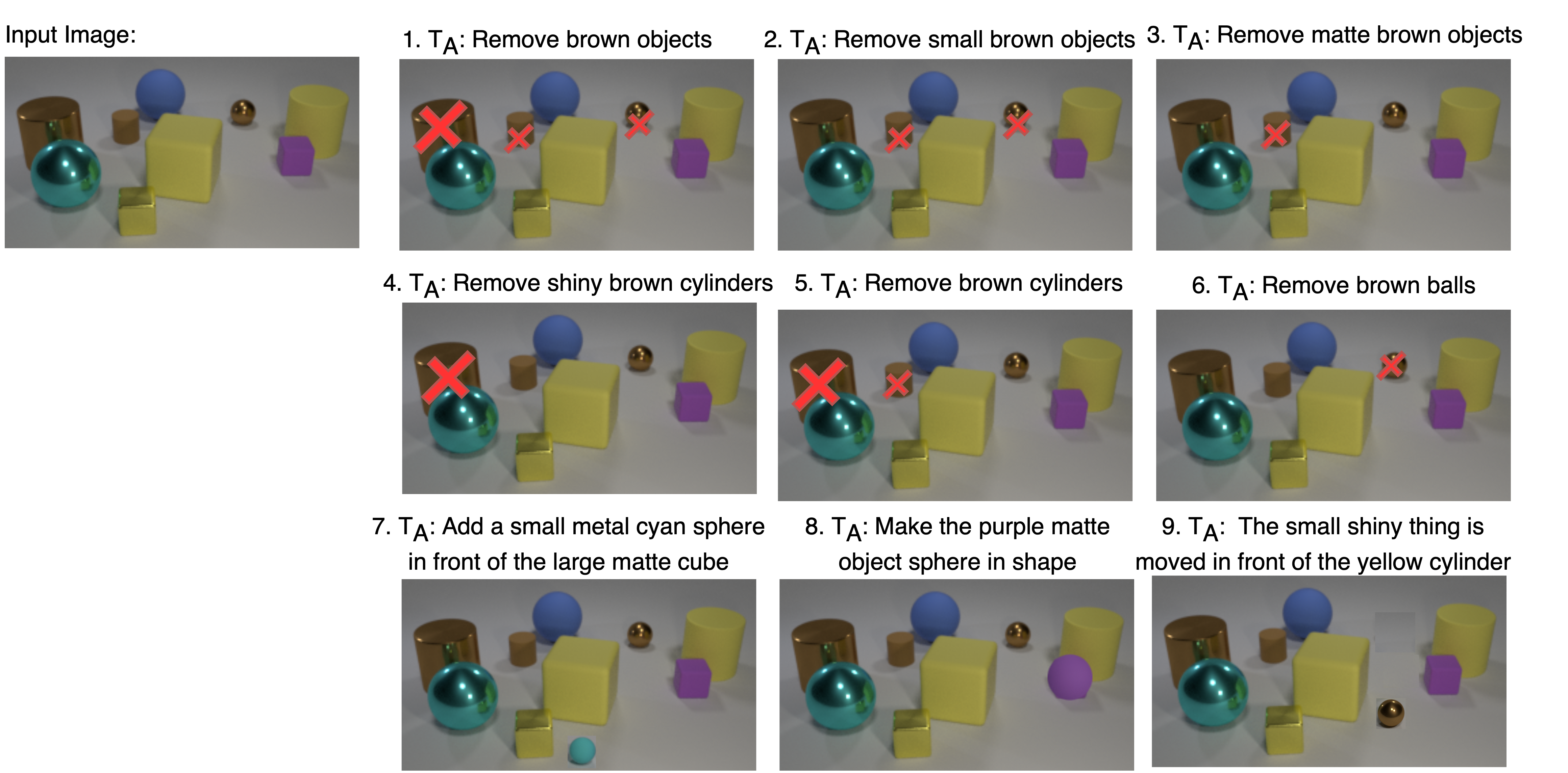} 
  \caption{Correct scene graph predictions for the given (input image, action text) by our ARL model}
  \label{fig:goodex}
\end{figure*}

Though understanding changes caused by actions is the core challenge in CLEVR\_HYP \cite{sampat2021clevr_hyp} task, it has a question answering downstream task. To answer questions in the CLEVR\_HYP dataset, models should be able to perform counting, check the existence of objects given the criteria, compare sets of objects, or retrieve attributes of the desired objects. We carry out a similar analysis of models based on their capability to perform above-mentioned reasoning tasks. The results are summarized in Table \ref{tab:qhtype}. 


For validation set, our proposed method ARL has better fine-grained accuracy across all reasoning types, but improvements on `query attribute' and `exist' types are maximum (6.8\% and 4.7\% respectively). For the 2HopQ$_H$ test set, the model is expected to perform logical operations within a particular reasoning type. For example, `How many objects are either red or cylinder?' and `Are there any rubber cubes that are not green?'. Though we see some gains here as well, our overall pipeline is limited by the capabilities of the question answering model of \citet{yi2018neural} we use in stage-3.

\subsection{Qualitative results} 

In Figure \ref{fig:goodex}, we visually demonstrate scene graphs predicted by our ARL model over a variety of action texts. From examples 1-6, we can observe that the model can correctly identify objects that match the object attributes (color, size, shape, material) provided in the action text. Examples 4 and 6 demonstrate that our system is consistent in predictions when we use synonyms of various words (e.g. sphere$\sim$ball, shiny$\sim$metallic) in the dataset. Finally, examples 7-9 show that our model does  reasonably well on other actions (add, change, move).

We further generate a t-SNE plot of action vectors learned by our best proposed model, which is shown in Figure \ref{fig:awesome_image1}. At a first glance, we can say that the learned action representations formulate well-defined and separable clusters corresponding to each action type. Clusters for add, remove and change actions are closer and somewhat overlapping. We observed that many samples of type `change' is  interpreted by the reasoner as `remove+add' action. For example, if a color of 'small blue metal sphere' is changed to 'red', the action reasoner interprets it as removal of the 'small blue metal sphere' followed by an addition of a 'small red metal sphere' on the same location. 


\subsection{Ablations}
\label{sec:abl}

\paragraph{Importance of stage-1 training}
Cause-effect learning with respect to actions is a key focus in CLEVR\_HYP. In existing models, it is formulated as a updated scene graph prediction task (i.e. given an initial scene and an action, determine what the resulting scene would look like after executing the action). In our opinion, stage-1 plays a critical role in learning causal structure of the world. To demonstrate this, we set up two experiments; first, where training takes place in a sequential manner (stage-1 followed by stage-2), where trained encoder-decoders from stage-1 are frozen and utilized in stage-2. Second experiment, where there is no separate stage-1 training and encoder-decoder in stage-2 are randomly initialized. 

\begin{figure*}
\minipage{0.56\textwidth}
  \includegraphics[width=0.7\linewidth]{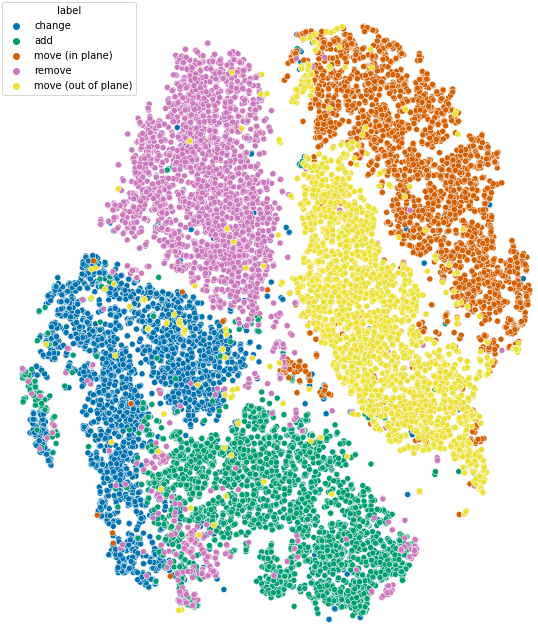}
  \centering
  \caption{The t-SNE plot of learned action vectors}\label{fig:awesome_image1}
\endminipage\hfill
\minipage{0.45\textwidth}
  \centering
  \includegraphics[width=0.8\linewidth]{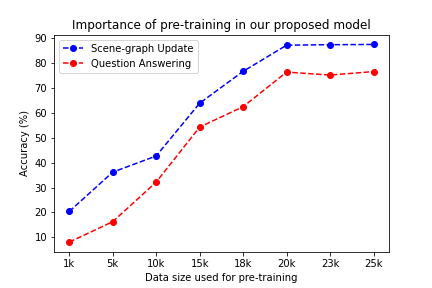}
 
  \includegraphics[width=0.8\linewidth]{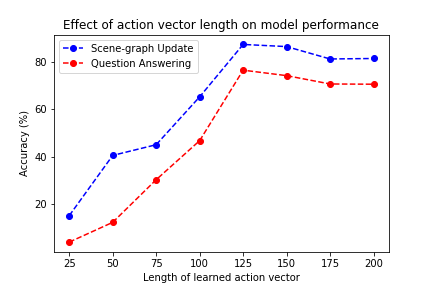}
  \caption{Performance of our model with varying (top) data size and (bottom) action vector lengths}
  \label{fig:graph1}
\endminipage \hfill
\end{figure*}

The results are summarized in Table \ref{tab:abl1}. We can observe that inclusion of stage-1 training improves the accuracy of scene graph prediction by $\sim$30\% compared to the stage(2 only) model. To evaluate question answering task of CLEVR\_HYP, both setups are followed by stage-3 where the image parser and scene-graph question answering modules are combined to predict the answer. It is known that \cite{yi2018neural} has near-perfect performance on the scene graph question answering task over CLEVR \cite{johnson2017clevr}. As a result, the gains achieved in the scene graph task directly benefit the question answering performance without much of a loss. In other words, there are only  0.2\% instances where the scene prediction is correct but the final answer is incorrect.

\paragraph{Performance with different lengths of learned action vector in stage-1}
In this ablation, the goal is to find out optimal length of action vectors that can reasonably simulate the effects of the actions. We experiment with different lengths of learned action vector- from 25 to 200 in increment of 25. Figure \ref{fig:graph1} (bottom) shows the effect of training with diverse action vector lengths on scene graph update and downstream question answering task. The model learns better initially when the vector length is increased, however performance reaches at peak for the action vector length of 125. 

\begin{table}[h!]
\begin{tabular}{@{}ccc@{}}
\toprule
\multicolumn{1}{l}{\textbf{Task}}                                              & \textbf{Experiment} & \textbf{Accuracy (\%)} \\ \midrule
\multirow{2}{*}{\begin{tabular}[c]{@{}c@{}}Scene Graph \\ Update\end{tabular}} & Stage(2 only)       & 56.3                   \\
                                                                               & Stage(1+2)          & 87.2                   \\ \midrule
\multirow{2}{*}{\begin{tabular}[c]{@{}c@{}}Question \\ Answering\end{tabular}} & Stage(2+3 only)     & 45.7                   \\
                                                                               & Stage(1+2+3)        & 76.4                   \\ \bottomrule
\end{tabular}
\caption{Performance of our model in the absence and presence of stage-1 over ordinary test set}
\label{tab:abl1}
\end{table}

\section{Related Works}

In this section, we discuss existing research efforts that align with our work in this paper. Specifically, we elaborate on tasks involving learning action representations, counterfactual reasoning, and what-if question answering datasets.

\paragraph{Tasks involving learning representation of actions:} Better representation learning is key to success in all kinds of artificial intelligence problems \cite{Banerjee_2021_ICCV, NEURIPS2020_49562478, chen21_interspeech,lee2018spoken, lee2018odsqa}. Learning a mapping from the goal (provided in natural language) to a sequence of actions to be performed in a visual environment is a common task in robotics \cite{kanu2020following,ALFRED20}. Specifically, human-in-the-loop methods for training robots to perform various actions involve learning a mapping between verbal commands and low-level motor controls of a robot \cite{stepputtis2020language}. Another relevant task is vision-and-language navigation \cite{mattersim,chen2018touchdown,nguyen2019vnla}, where an agent navigates in a visual environment to find the goal location by following natural language instructions. Navigation tasks focus on selecting the right actions to achieve desired goals provided a visual environment and natural language instructions. Our focus in this paper is to develop models that can implicitly reason about the effect of actions rather than determining which action to perform. 

\paragraph{\textbf{Counterfactual vision-language reasoning:}} 
Counterfactual reasoning is termed as an ability to develop mental representations to generate alternate consequences about an event that happened in the past based on given criteria. Inspired by this human ability, there have been efforts to utilize this concept to improve many aspects of language and vision-language research;  \citet{kusner2017counterfactual} and \citet{garg2019counterfactual} proposed methods to measure counterfactual fairness of models. A few recent works incorporated counterfactual augmentation of training sets  \cite{zmigrod2019counterfactual, fu2020counterfactual} to improve the robustness of models and discourage biases. Contrary to that, the work of \citet{fu2020iterative} was the first to utilize counterfactual instructions in training (ex. multiple questions asked to the same image set) to deal with the data scarcity issue and improve the generalization.

\paragraph{\textbf{What-if question answering datasets:}} 
WIQA \cite{tandon2019wiqa} is a testbed for what-if reasoning over natural language contexts. Provided a procedural paragraph, the task is to answer the question ``Does change in X result in change in Y?'' (where X and Y are two events from the paragraph) as a 3-way choice- correct, opposite, or no effect. In the vision-language domain, TIWIQ \cite{wagner2018answering} was among the earliest works. 
Given a synthetically rendered table-top scene, the task is to generate a textual response to the what-if question when an action (push, rotate, remove or drop) is performed on an object. 
However, the evaluation of open-ended text generation is challenging. 
To fill in this gap, \citet{sampat2021clevr_hyp} created CLEVR\_HYP dataset. It shares similarities with TIWIQ for having rendered images, limited action types, and QA as a task. However, the key difference is that in CLEVR\_HYP, an action can cause changes to multiple objects in the scene, which is not the case with TIWIQ. 

\section{Conclusion}

In the vision and language domain, several tasks are proposed that require an understanding of the causal structure of the world. In this work, we propose an effective way of learning action representations and implement a 3-stage model for the what-if vision-language reasoning task CLEVR\_HYP. We provide insights on the learned action representations and validate the effectiveness of our proposed method through ablations. Finally, we demonstrate that our proposed method outperforms existing baselines while being data-efficient and showing some degree of generalization capability. By extending our approach to a larger set of actions, we aim to develop AI agents which are equipped with action-effect reasoning capability and can better collaborate with humans in the physical world. 

\section*{Limitations}

 In the CLEVR\_HYP dataset, all actions are considered to be independent of each other and execution of actions is always guaranteed. However, we found few instances in the CLEVR\_HYP dataset where two different actions taken over an initial scene leads to the same resulting effects. Our proposed 3-stage model has higher error rates and low confidence for such samples. Further, in real world situations, actions can have inter-dependencies on world conditions or have other properties such as order-sensitivity, symmetry with respect to other actions, reversibility etc. Exploring hypothetical reasoning problems from aforementioned aspects is still an open research direction. 
 
 \section*{Computing Infrastructure} All experiments are done over Tesla V100-PCIE-16GB GPU. Total time for all experiments (including parameter search for best model) utilized approximately 70 GPU hours. 

\section*{Ethical Considerations} In this paper, our experiments are limited to publicly available CLEVR\_HYP dataset that is synthetically generated through controlled environment. Thus, there are no ethical violations or known bias issues, to our best knowledge.

\section*{Acknowledgements} 
We are thankful to the anonymous reviewers for the constructive feedback. This work is partially supported by the grants NSF 1816039 and NSF 2132724.

\bibliography{emnlp2022}
\bibliographystyle{emnlp2022}

\appendix

\end{document}